# Modeling the Developmental Shift in Telicity Acquisition

**Ellie Xia**[1] **Parisa Kordjamshidi**[1] **Alan Hezao Ke**[2]
[1]Michigan State University [2]University of South Carolina
{xiayili, kordjams}@msu.edu hezao@mailbox.sc.edu

## Abstract

Acquiring telicity, which is the distinction between bounded (e.g., *ate an apple*) and unbounded (e.g., *ate apples*) events, requires first language (L1) learners to map surface-level and semantic cues to abstract event structures, but the computational trajectory of this mapping is not well understood. We introduce a **Difference in Surprisal** method that uses GPT2 token surprisal over paired temporal adverbial diagnostics (*in an hour* versus *for an hour*) to automatically label telicity across English CHILDES corpora, validated against expert linguist judgments. Using these labels, we train diagnostic logistic regression classifiers on 12 syntactic and lexical semantic features to compare how child speech and child-directed speech encode telicity. The two models diverge: the child model reaches near perfect accuracy through a single deterministic cue, the presence of a post-verbal determiner, while the adult model relies more heavily on verb class and other lexical semantic features, with the determiner cue neutralized. This trajectory supports Syntactic Bootstrapping: learners first exploit high-frequency structural cues as a scaffold to bootstrap, before developing fully compositional, verb-based event structures.

## 1 Introduction

Telicity is a core concept in the event structure of predicates, which distinguishes whether an event is bounded and culminates ('telic') or is characterized by ongoing duration without an inherent endpoint ('atelic') (Levin and Rappaport Hovav, 1994; Krifka, 1998; MacDonald, 2008; Portelance et al., 2023; Kim et al., 2024). For language learners, this distinction is crucial yet complex. Infants must learn to map surface-level morphosyntactic, syntactic, and semantic features, such as determiners, resultative particles, and verb classes, to their corresponding aspectual semantics (Wagner, 2006). Telicity is formally diagnosed by the compatibility of predicates with temporal adverbials (Dowty, 1979; Smith, 1991; MacDonald, 2010). More precisely, "in an hour" phrases are semantically felicitous with telic predicates as they denote the time span of culmination, whereas "for an hour" phrases select atelic predicates, expressing duration. For example, Figure 1 illustrates the culmination point of the event *eating an apple* is where the final state of *only the core remaining* serves as the telic boundary. The incremental stages, such as *one bite*, *half eaten* are intermediate process leading up to this final change of state.

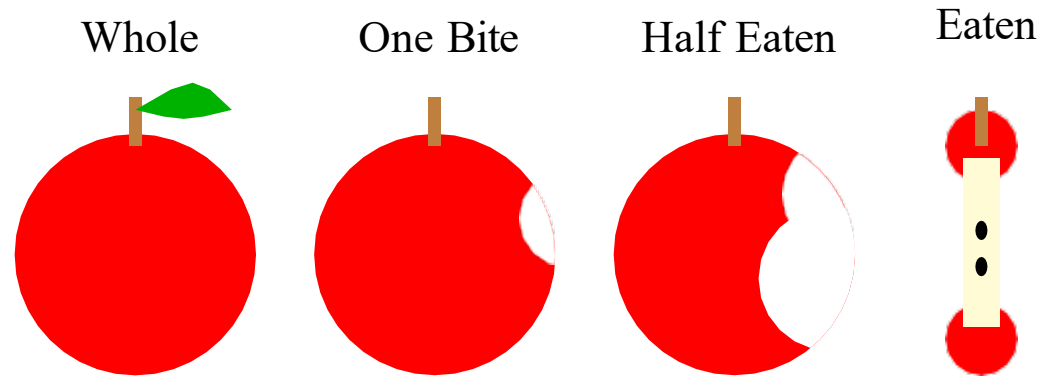

Figure 1: Apple Incremental Consumption

To give a concrete natural language example, consider the following contrast in (1) when the sentence is appended with different temporal phrases:

(1) I ate **an** apple {in an hour / #for an hour}.

Example (1) is not compatible with "for an hour" because the event *ate an apple* is terminated/completed when *an apple* is being eaten. Thus "in an hour" captures the termination of the event but "for an hour" does not. The symbol # indicates semantic anomaly or infelicity for appending "for an hour". This means that although the sentence follows the rules of grammar, the meaning sounds infelicitous or odd in that context.

More telicity related cues including plurality (2), resultative (3), partial (4), aspectual (5), and stative (6) readings, are as follows:

(2) I ate apple**s** {#in an hour / for an hour}.

(3) I ate **up** all the apples {in an hour / #for an

hour}.

(4) I ate **a piece of** the apple {in an hour / #for an hour}.

(5) I **finished** eating all the apples {in an hour / #for an hour}.

(6) I **like** eating apples {#in an hour / for an hour}.

Since these distinctions (1)-(6) rely on the interface between syntax (determiner, Structural Case (Chomsky, 1957)) and semantics (boundedness, verb class), acquiring them requires L1 learners to track complex distributional cues.

We investigate the acquisition of these different cues in early English-speaking children by connecting linguistic theory with computational modeling. Specifically, we test the following hypotheses regarding the developmental trajectory of telicity. These hypotheses are potentially compatible with one another and are not necessarily mutually exclusive:

**$H_1$**: Telicity is based on the boundedness of the object (semantics of NP) (Schmitt, 1996; Krifka, 1998; MacDonald, 2010; Xia and Ke, 2026).

**$H_2$**: Telicity is primarily a property of the verb. (Vendler, 1967; Verkuyl, 1972; Dowty, 1979; Naigles, 1990; Wagner, 2006; Ramchand, 2008).

**$H_3$**: Telicity is induced by prepositional phrases or resultative particles (e.g., `up`, `off` in English) (Cheng and Huang, 1994; Sybesma, 1997; Lin, 2003; Kratzer, 2005; Song, 2018; Tay, 2024).

**$H_4$**: Telicity is a mixed semantic property of different cues (Borer, 2005; Zhao et al., 2021; Hu, 2024).

The current study addresses L1 acquisition on telicity, specifically targeting the gap between formal linguistic theories of event boundedness and the empirical distribution of telicity cues in naturalistic child and child-directed speech by applying natural language processing methods, based on CHILDES data (MacWhinney, 2000) and recent developments in large language models (LLM) (Radford et al., 2019). Using LLM-based token surprisal and probability-based analysis within a computational pipeline, this project tracks how the use of telic predicates by children develops over time compared to adults' speech. The results will inform theories of the acquisition of aspectual semantics by highlighting developmental patterns and potential avoidance behaviors, demonstrating how computational linguistics can enrich traditional acquisition research. The pipeline is shown in Figure 2. It consists of three primary stages designed to test our core hypotheses ($H_1$ to $H_4$). First, in Stage I, we partition the English CHILDES corpora into child utterances and adult child-directed speech. Second, in Stage II, we extract 12 features, while an LLM-based Labeling Engine calculates the diagnostic surprisal difference ($\Delta S$) to determine the telicity of each utterance. Finally, Stage III applies diagnostic probing via logistic regression. By comparing the feature weights relied upon by the child model versus the adult model, this framework allows us to quantitatively track the developmental shift from syntactic bootstrapping to semantic composition.

In framing these hypotheses, we specifically evaluate the mechanism of the Syntactic Bootstrapping Hypothesis (Gleitman, 1990) under the learnability constraints of the Transparency Principle (Wagner, 2006; Lightfoot, 2017), which posits that learners initially prioritize transparent, one-to-one surface mappings before mastering complex compositional operations. Furthermore, we clarify that our empirical scope targets language *production* via naturalistic corpus transcripts rather than experimental comprehension. While experimental studies probing sentence acceptability test comprehension, observational corpus studies reflect production. Prior acquisition literature notes a marked asymmetry where children may demonstrate adult-like competence in comprehension while lagging in production, or vice versa (Wagner, 2001; Conroy and Lidz, 2007; Hendriks, 2016).

**Contributions**: **(a)** Introduce a novel "Difference in Surprisal" method to automatically annotate telicity in large corpora using open-weight LLM. **(b)** Provide quantitative computational evidence supporting the Syntactic Bootstrapping Hypothesis in early L1 acquisition. **(c)** Identify a measurable developmental shift where reliance on syntactic cues (determiners) transitions to lexical semantic cues (verb class).

## 2 Related Work

Telicity lies at the interface of the verb class, morphology, and syntax, etc. Prior research has examined this interface through two lenses: developmental psycholinguistics and computational modeling (Bloom et al., 1980; Shirai and Andersen, 1995; van Hout, 1998; Zhao et al., 2021; Kim et al., 2024).

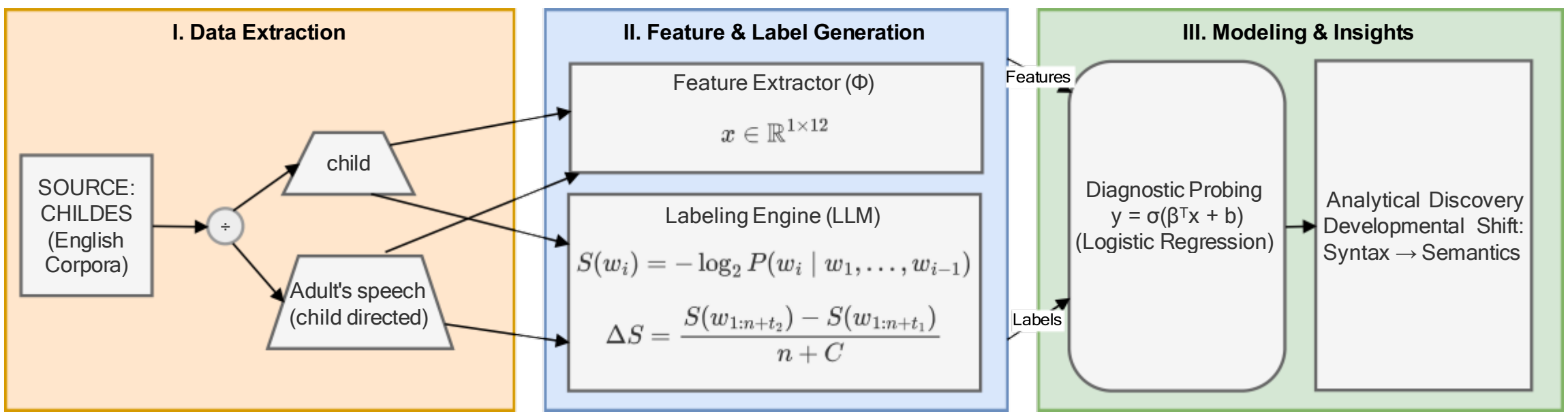

Figure 2: L1 Telicity Acquisition Pipeline

### 2.1 Theoretical Acquisition of Telicity

Theoretical work suggests that children acquire aspectual mappings in a non-uniform manner, heavily influenced by the **Transparency Principle** (Wagner, 2006; Lightfoot, 2017). This principle predicts that one-to-one form-meaning mappings are acquired earlier than complex compositional ones.

**Overt vs. Compositional Cues.** Children as young as age 3 reliably interpret verb-particle constructions (e.g., *up* in English, *op* in Dutch) as encoding telic events (Wagner, 2006). These particles serve as transparent markers of completion. In contrast, children show a marked delay in acquiring telicity induced by object quantization (e.g., *drink milk* vs. *drink a glass of milk*). Adult-like judgments for these compositional structures often do not emerge until after age 5 (van Hout, 1998).

**Tense-aspect Morphology Selection.** Early production data reveals that young children initially restrict tense-aspect morphology to specific semantic classes. Bloom et al. (1980) and Shirai and Andersen (1995) observe that children predominantly apply past-tense marking to telic verbs (e.g., *broke*, *made*) and progressive marking to atelic verbs (e.g., *riding*, *laughing*). This suggests that early grammar is heavily lexicalized before becoming fully productive.

**Cross-Linguistic Variation.** The strength of these mappings varies by languages. Dutch adults, for instance, assign telic interpretations to bounded objects far more consistently ($\approx$ 78%) than English adults ($\approx$ 25%), suggesting that English relies more heavily on pragmatic inference than on object quantization alone in the experimental work (van Hout, 1998).

### 2.2 Computational Modeling of Telicity

Recent computational work has begun to probe whether neural networks capture the structural and lexical nuances required to model aspectual interpretation and clause-level situation types (Siegel and McKeown, 2000; Zarcone and Lenci, 2008; Zhao et al., 2021; Friedrich et al., 2023; Kim et al., 2024).

**Automatic Aspectual Classification.** Early machine learning frameworks approached aspectual detection as a supervised task optimized over type-level linguistic markers. The foundational work of Siegel and McKeown (2000) introduced methods to combine sparse, rule-based linguistic indicators–such as the frequency of progressive markers or temporal adjuncts–to classify whether a verb root function natively behaves as stative or dynamic. Building upon these type-based baselines, Friedrich et al. (2016) extended the task to the token-clausal level, using sequential labeling to capture clause-level situation entities (e.g., episodic events, habits, or states). More recently, Friedrich et al. (2023) provided a survey mapping these historical feature-engineering baselines to modern contextualized spaces, emphasizing that tracking compositional aspect remains a critical, non-trivial bottleneck for modern NLP.

**Distributional Semantics.** To relax the need for hand-crafted syntactic rules, Kim et al. (2024) explored aspectual boundaries through distributional semantics. Using GloVe embeddings, they modeled telicity as a function of lexical vector co-occurrence, testing whether the raw statistical probability of structural complements appearing alongside specific verbal heads could predict human telicity judgments. Their findings demonstrate that while static distributional signals correlate with coarse telic environments, they inherently lack the hierarchical structural dimensions necessary to compute complex compositional telicity or capture fluid aspectual boundaries in context.

**Aspectual Composition and Coercion.** Another line of structured computational research evalu-ates semantic coercion—where an inherently atelic predicate is forced into a bounded interpretation by its arguments. Structured distributional models and distributional memory frameworks (Chersoni et al., 2019) have been deployed to quantify the cognitive processing load of these aspectual transitions. These models show that tracking thematic fit and argument-driven expectations can capture semantic composition, though they often sctruggle to scale outside strictly constrained, experimental sentences.

**Transformer Sensitivity.** Most recently, researchers have turned to deep, pretrained language models to probe implicit semantic boundaries. Zhao et al. (2021) demonstrated that pretrained transformers (e.g., BERT, RoBERTa) can infer certain telicity patterns, particularly those anchored to explicit lexical cues. However, their findings reveal that models do not fully match human sensitivity; while humans integrate a wide range of pragmatic, flexible, and contextual cues, transformers rely disproportionately on shallow, surface-level artifacts like post-verbal determiners or explicit temporal units.

## 3 Methodology

Our approach combines LLM-based surprisal with classical machine learning (i.e., linear regression model) to understand how telicity cues are used by the learners. First, we calculate the difference in token surprisal between telic and atelic diagnostic contexts to automatically generate binary telicity labels across the corpus. Second, using these LLM-derived annotations as our target labels, we train diagnostic classifiers to determine which linguistic features best predict telicity labels. Last, by comparing the feature weights of a model trained on child speech against one trained on adult child-directed speech, we can quantitatively map the developmental trajectory of these cues. The details of this process are explained in the following sections.

### 3.1 Data and Preprocessing

To investigate how L1 speakers, specifically English speaking children, learn the concept of *telicity*, we utilize the CHILDES database (MacWhinney, 2000). In particular, we extract English corpora including English-NA (North American) and English-UK (British) (MacWhinney, 2010). We partition the data into Child Speech (utterances by the child) and Parent/Caregiver(Adult) Speech (utterances directed at the child). To process the raw corpus transcripts for computational modeling, we first clean the data by removing non-linguistic transcriber annotations (e.g., pause markers, phonetic notes). Utterances are filtered to exclude extreme length outliers. Short fragments under three tokens are removed. Utterances that do not contain a verb are removed because they lack the core predicate-argument that is necessary, given that adjectival predicate is always stative (atelic). Conversely, utterances exceeding 30 tokens are excluded to prevent transcription artifacts from dependency parsers (de Marneffe et al., 2021). This yields a dataset optimized for information-theoretic evaluation. The processed datasets are used to extract the 12 linguistic features described in Section 3.3. The use of adult child-directed speech (CDS) as the adult benchmark is grounded in extensive developmental literature demonstrating that CDS input constitutes the primary linguistic environment shaping L1 distribution and constraint induction (Snow, 1972; Farwell et al., 1979). The full filtering breakdown—from raw counts to post-filtering totals across subsets—is detailed in Appendix.

### 3.2 Telicity Labels via LLM-Based Surprisal

To generate consistent telicity annotations in a large corpus without manual effort, we propose a novel method, namely the "Difference in Surprisal" method. Previous research has established that LLM-based surprisal serves as a robust diagnostic for recovering latent grammatical structure and identifying semantic infelicity (Kwiatkowski et al., 2012; Goodkind and Bicknell, 2018; Fan and Reilly, 2020; Wilcox et al., 2020; Cong, 2024; Gambi et al., 2024). We define telicity based on the compatibility with temporal adverbials: *telic* predicates prefer $t_1$ ="in an hour" (bounded), while *atelic* predicates prefer $t_2$ ="for an hour" (durative).

We employ LLM (GPT-2) to calculate the token-level surprisal $S(w_i)$. The surprisal of a token is defined as the negative log probability of its occurrence given the context:

$$S(w_i) = -\log_2 P(w_i \mid w_1, \ldots, w_{i-1}) \qquad (1)$$

For a given sentence context (e.g., "He ate the apple"), we append the diagnostic phrases and compute the mean surprisal over the total phrase tokens ($n + C$= the number of tokens in the original sentence plus the three-token diagnosis phrase "for an hour" or "in an hour"). The diagnostic signal $\Delta S$ is defined as:

$$\Delta S = \frac{S(w_{1:n+t_2}) - S(w_{1:n+t_1})}{n + C} \quad (2)$$

A positive $\Delta S$ indicates that sentence with "for an hour" is more surprising (less likely) than sentence with "in an hour," signaling a more **Telic** event, and the bigger the positive number, the more likely the sentence is telic. We discretize this continuous score into binary labels (Telic/Atelic) by applying the default threshold teasing apart telic from atelic events.

Crucially, rather than prompting the language model to generate categorical judgments (Zheng et al., 2023), our Difference in Surprisal method derives binary labels through next-token probability aggregation over constrained temporal modifiers. Using mean token surprisal (normalizing the summed negative log-probabilities by phrase length $n + C$) ensures that the metric captures per-token information density without introducing length-dependent bias.

### 3.3 Features for Telicity Classification

We extracted the 12 linguistic features using an automated Python pipeline. Tokenization, lemmatization, and POS tagging, which identify morphosyntactic cues, were performed with Stanza (Qi et al., 2020). Syntactic configurations, including core arguments and linear dependencies like verb_det_dist, were mapped using the Universal Dependencies parser in Stanza. Finally, we operationalized Lexical Semantics (Verb_Class) by matching lemmatized verbs against a custom, manually annotated aspectual verbs.

**Syntactic Configuration:** These features capture the internal structure and complexity of the VP, specifically looking for cues of boundedness. This analysis includes binary and count-based measures of the words that follow a verb. First, this involves checking whether or not determiners like 'the' or 'a' appear immediately after the verb (`n_determiner_after_verb`) (Xu and Schmitt, 2025a,b,c), the number of particles like 'up' or 'off' (`n_particle`), and checking for the presence of prepositional phrases that indicate a destination or result (`has_PP_goal`). Second, we also include linear distance metrics which measure the gap between the verb and its introductory word (`verb_det_dist`) or the verb and its actual object (`verb_obj_dist`), both of which serve as proxies for the syntactic configuration of the predicate (Gibson, 2000; Liu et al., 2017; Futrell et al., 2015; Levy, 2008; Gildea and Temperley, 2010). Third, we add the presence of core arguments (`n_subject`, `n_obj`) which are necessary to saturate the predicate's thematic requirements

**Lexical Semantics:** We incorporate the inherent aspectual class of the verb (`Verb_Class`), based on the classification of argument structure (Chomsky, 1957, 1995; Borer, 1993; Levin and Rappaport Hovav, 1994; Culicover and Jackendoff, 2005; Ramchand, 2008; Roberts, 2010). This feature identifies whether the verb is inherently telic or not, which provides the semantic baseline for telicity judgments (Anderson, 2017). We characterize Verb_Class into three types: `Verb_Class_0` (Unergative): Intransitive verbs where the subject is an agent (e.g., *cry*, *dance*). These are inherently atelic as they describe ongoing activities without a natural endpoint. `Verb_Class_1` (Unaccusative): Intransitive verbs where the subject undergoes a change of state (e.g., *come*, *arrive*, *fall*). These are inherently telic (eventive), as the endpoint is built into the lexical meaning. `Verb_Class_2` (Transitive): Verbs that take a direct object (e.g., *eat*, *drink*). These exhibit variable telicity (aspectual composition), where the boundedness of the event is determined by the quantization of the internal argument (e.g., "drinking milk" vs. "drinking a glass of milk").

**Morphosyntactic Cues:** These features track overt grammatical aspect. We operationalize verb particles via a binary presence indicator `has_particle` (replacing count metrics to avoid spurious multiverbal readings). To capture aspectual morphology without redundancy, we distinguish between `has_aspect_marker` (a binary flag denoting the presence of an auxiliary such as perfect *have/has*) and `aspect_marker_count` (an integer feature capturing stacked aspectual marking, such as perfect progressive *have been V-ing*). In English, these stacked aspectual layers fundamentally alter telic boundedness relative to simplex aspectual marking.

### 3.4 Diagnostic Probing Models

To compare the "acquisition logic" of children versus adults, we train **Logistic Regression** classifiers to predict the LLM-derived telicity labels (as described in Section 3.2) using the 12 extracted features. We train two separate models: Child Model is trained on child's utterances; Parent Model is trained on adult's utterances.

By analyzing the coefficients ($\beta$) of these logistic regression models, we quantify the importance of each linguistic cue. A high positive coefficient for a feature (e.g., `n_determiner_after_verb`) implies that the presence of the feature is a strong predictor of telicity for that group. This allows us to test our hypotheses ($H_1$ to $H_4$) by observing which features impact the decision boundaries on telicity labels.

## 4 Results

Looking at the experimental results in Table 1, we can go back to evaluate the hypotheses and explain how the presence of post-verbal determiners as part of syntactic configuration directly informs producing telicity by children. We first analyze the performance of the diagnostic probing models (Logistic Regression) trained to predict the LLM-derived telicity labels from linguistic features in Adult vs. Child speech. We apply 70% of our data for training and 30% for testing.

**Child Model:** In contrast, the Child Model (Table 1) achieves perfect convergence (100%) across all metrics. As it will be discussed in Section 6, this is not a result of superior generalization, but rather indicative of a strong categorical separation driven by a single feature (`n_determiner_after_verb`). The model effectively learns a deterministic rule: the presence of a determiner implies telicity in the child subset.

**Adult Model:** The Adult model is trained on the adult speech. As shown in Table 1, the model achieves a robust performance with an accuracy of 0.93 and a macro F1-score of 0.84 in the testing phase.

Table 1: Performance metrics for Child and Adult regression models based on 12 linguistic features

| Model | Class | Metric | Train | Test |
|---|---|---|---|---|
| Child | Atelic | Precision | 1.00 | 1.00 |
| | | Recall | 1.00 | 1.00 |
| | | F1-Score | 1.00 | 1.00 |
| | Telic | Precision | 1.00 | 1.00 |
| | | Recall | 1.00 | 1.00 |
| | | F1-Score | 1.00 | 1.00 |
| | **Overall** | **Accuracy** | 1.00 | **1.00** |
| Adult | Atelic | Precision | 0.71 | 0.94 |
| | | Recall | 0.84 | 0.98 |
| | | F1-Score | 0.77 | 0.96 |
| | Telic | Precision | 0.80 | 0.73 |
| | | Recall | 0.66 | 0.48 |
| | | F1-Score | 0.72 | 0.58 |
| | **Overall** | **Accuracy** | 0.75 | **0.93** |

## 5 Expert Validation

To validate that our LLM-derived surprisal metric (`Diff_MeanBits`) aligns with human intuition, we evaluated the metric against human-annotated sentences. We conducted an expert judgment task. Three American English-speaking linguistic Ph.D. or Ph.D. students evaluated a stratified sample of 200 utterances ($n = 100$ Child, $n = 100$ Child-Directed). Linguists judged whether each sentence was more compatible with the telic diagnostic *in an hour* or the atelic diagnostic *for an hour*, or both if examples can have a two-way reading, which means the sentence can have both a telic and an atelic reading. Let $P_1$ and $P_2$ denote the proportions of telic and atelic judgments, respectively, computed over all three votes from the experts. We calculated a Telicity Index ($TI$):

$$TI = \frac{P_1}{P_1 + P_2} \quad (3)$$

An utterance is assigned a Ground Truth of **Telic** if $TI >= 0.5$, and **Atelic** otherwise. Then we compare the labels produced by surprisal difference in 3.2 with the experts' annotation.

We first use raw surprisal differences with a naive threshold of **0**. The metric captures a highly predictive signal for human judgments (Pearson $r = 0.266$), yielding an initial classification accuracy of **70.5%**. Notably, this raw correlation performs competitively alongside established evaluation benchmarks for human annotation (Pearson $r = 0.217, p < 0.001$; Peelle et al. 2020). In Table 2, the pre-threshold confusion matrix reveals a conservative bias: the model is accurate at identifying atelic events (True Negatives = 120) but underestimates telic events.

Table 2: Confusion Matrix (Pre-Threshold).

| | Pred Atelic | Pred Telic |
|---|---|---|
| **True Atelic** | 120 | 34 |
| **True Telic** | 25 | 21 |

### 5.1 Inter-Rater Reliability and Human Consensus

The evaluation data was independently coded by three expert linguists. Inter-rater reliability metrics demonstrate highly robust convergence and a stable semantic signal across the dataset. The annotators achieved a majority agreement rate (≥ 2/3 raters) of 96.00% (192/200 sentences), corroborating the ≈95.5% reliability baseline observed in preliminary development cycles. Absolute unanimity (3/3 raters) was observed in 43.50% (87/200) of the test items.

While absolute unanimity (3/3) reached 43.50%, this variation does not signify noisy annotation quality, but reflects the inherent semantic indeterminacy of aspectual interpretation. As observed in recent computational work on telicity by Ma and Miyao (2026), there exists a persistent divergence between discrete theoretical categories and empirical human perception, claiming that telicity judgments inherently cause gradient acceptability. In our evaluation sample, disagreements predominantly stemmed from ambiguity flags (69/200 items where raters permitted dual readings or split on optional contextual licensing). These ambiguous items split evenly across child ($n = 39$) and parent ($n = 30$) speech and were symmetrically distributed across determiner-present and determiner-absent clauses, confirming that rater variance is not due to the post-verbal determiner effect. Furthermore, rater divergence also showed degrees of pragmatic accommodation: stricter syntactic readings disallow coercion in sentences like *I ate an apple for an hour*, whereas more permissive readers accommodate partial consumption readings (Krifka, 1998; Xu, 2026).

As summarized in Table 3, the model exhibits alignment with human linguistic judgments. The model achieved a lenient accuracy rate of 87.50% against the consolidated human majority consensus, with an average per-rater accuracy of 88.50%.

Table 3: LLM Agreement and Accuracy Rates Under Evaluation from three LRs (Linguist rater) ($N = 200$)

| LR | Agree (%) | Sentence Count |
|---|---|---|
| #1 | 94.50% | 189 |
| #2 | 91.00% | 182 |
| #3 | 80.00% | 160 |
| **Average** | **88.50%** | *Cross-rater mean* |
| **Telicity Index** | **87.50%** | **175 / 200** |

## 5.2 Evaluation Threshold Optimization

We maintain an explicit distinction between the *labeling decision threshold* ($T = 0$) and the *empirical evaluation boundary* ($\epsilon$), which will be introduced in this section. The primary corpus labels are generated using a default threshold $T = 0$, which serves as the theoretically motivated point of indifference between the two diagnostic adverbials. Furthermore, in order to test if the default threshold 0 is aligned with human judgment, the sweep across candidate thresholds in Figure 3 represents a post hoc, descriptive validation designed to benchmark how continuous model surprisal aligns with human majority voting across a continuous Telicity Index. Optimizing this boundary yields an empirical cutoff of $\epsilon = -0.198$, which lies adjacent to the theoretical default of 0. Importantly, across both the child and adult corpora, over 98% of sentences fall strictly outside the narrow interval $[-0.198, 0)$, meaning that less than 2% of binary classifications would shift under the empirical boundary. We therefore preserve $T = 0$ for all primary experimental probing to prevent overfitting on the 200-sentence validation sample.

Table 4: Confusion Matrix (Post-Threshold).

| | Pred Atelic | Pred Telic |
|---|---|---|
| **True Atelic** | 49 | 28 |
| **True Telic** | 46 | 77 |

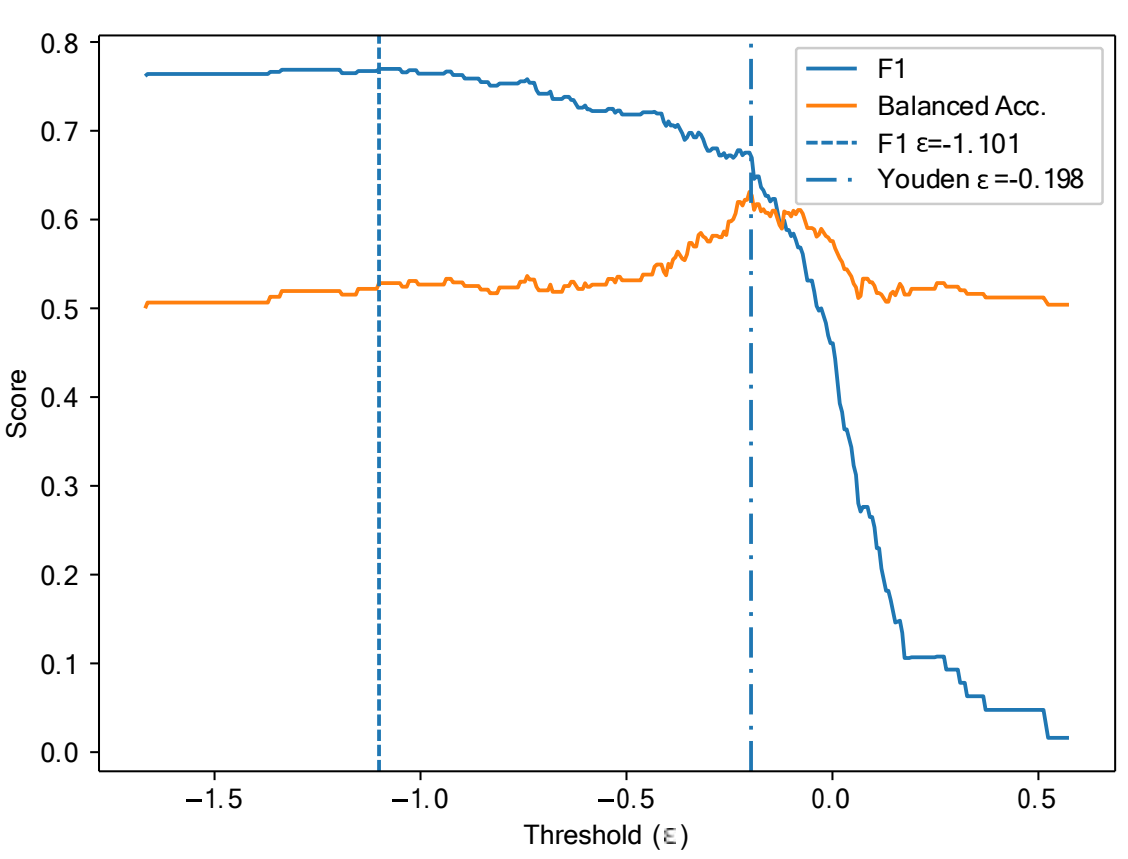

Figure 3: F1 score and balanced accuracy across thresholds for mean surprisal difference ($\Delta S$)

Under this empirical evaluation boundary $\epsilon = -0.198$, the point-biserial correlation between the binary model output and the continuous human expert index advances to $r = 0.2999$ ($2$ highly significant at $p < 0.00002$). The most pronounced statistical improvement occurs within the unambiguous categorical subset ($N = 53$), where items possess definitive human majorities (*strictly telic*

Table 5: Logistic Regression (selected) Coefficients in Child Model.

| Variable | Coefficient ($\beta$) | $z$-score | $P > \|z\|$ |
|---|---|---|---|
| **n_det_after_verb** | **16.07** | —$\dagger$ | *** |
| verb_det_dist | 3.7875 | 218.14 | *** |
| verb_obj_dist | 1.2410 | 115.88 | *** |
| n_subject | 0.7290 | 50.49 | *** |
| has_aspect_marker | 0.1141 | 2.44 | * |
| n_obj | -2.8449 | -146.21 | *** |
| verb_pos_ratio | -1.2218 | -119.44 | *** |

$\dagger$ *Note: Unstable* $z$*-score due to quasi-complete separation.*

Table 6: Logistic Regression (selected) Coefficients in Adult Model.

| Variable | Coefficient ($\beta$) | $z$-score | $P > \|z\|$ |
|---|---|---|---|
| Verb_Class_1 | 3.7520 | 2.99 | ** |
| verb_det_dist | 2.2804 | 24.75 | *** |
| Verb_Class_2 | 2.0871 | 11.82 | *** |
| verb_obj_dist | 1.2991 | 29.66 | *** |
| has_PP_goal | 0.5279 | 5.99 | *** |
| n_subject | 0.2379 | 4.64 | *** |
| **n_det_after_verb** | **-0.0210** | -0.22 | 0.821 (n.s.) |

or *strictly atelic*). Within this group, the categorical correlation rises from an unoptimized $r = 0.2666$ to an **optimized** $r = 0.4188$, achieving high statistical significance ($p = 0.0018$).

## 6 Discussion

### 6.1 "Determiner clue" in Child Speech

A finding in our results is the Child Model's near-perfect accuracy (1.00) and the extreme magnitude of the n_determiner_after_verb coefficient (16.07) compared to the Parent Model (−0.02). At first glance, this might suggest a trivial tautology in the dataset: telic events often contain determiners. However, the fact that this feature is the *sole* dominant predictor for children—but completely neutralized in the parent model—points to a reality that in parent speech, because parent utterances are syntactically more complex and lengthy, determiners are distributed more variably across the clause. Consequently, for adults, telicity appears to be derived via implicature rather than reliance on a specific structural cue.

Our results provide computational evidence for the Syntactic Bootstrapping Hypothesis (Gleitman, 1990; Wagner, 2006). The Syntactic Bootstrapping Hypothesis proposes that children learn the meanings of verbs by observing the grammatical structures (**the syntax**) in which those verbs appear. For example, if a child hears a verb with both a subject and an object, they can deduce it describes a relationship between two entities, even if they don't know the specific action yet (Wagner, 2006). In our experiment, child model essentially learns a heuristic: "If there is a determiner, the event is telic."

This finding directly supports the logic of bootstrapping as we have explained. If a simple linear regression model trained on child data can predict telicity based on the syntactic configuration of post-verbal determiners, it proves that these structural metrics are meaningful markers that a child can use to "bootstrap" their way into understanding complex telicity. Figure 4 indicates such result.

### 6.2 Structural cues: Distance Matters

The most striking finding from our logistic regression analysis (Table 5) is the outsized predictive power of verb_det_dist ($\beta = 3.78$) and the "perfect separation" caused by the presence of determiners. This suggests that explicit post-verbal syntactic configurations effectively override inherent lexical defaults during early language acquisition, aligning with classic structural bootstrapping accounts (Gleitman, 1990). Because determiners constitute a highly frequent, closed-class category, they provide a dense, reliable distributional cue in the immediate post-verbal domain (Gillette et al., 1999; Yang, 2016). In contrast, verbs belong to an open lexical class characterized by significantly higher structural variety and sparser individual token frequencies, making the localized phrasal geometry a more robust anchor for the model than underlying lexical aspect.

### 6.3 Developmental Shift: From Syntax to Semantics

The shift in predictive features suggests a fundamental change in telicity production. As shown in Table 6, the Adult model stands in substantial contrast to the Child Model (Table 5): the feature n_determiner_after_verb coefficient drops near zero for adults, while semantic features like Verb_Class become highly significant ($\beta = 3.752$). This divergence reflects a transition in the "acquisition trajectory" from syntactic to semantic cues, a cognitive shift visually summarized in Figure 4.

**Child Model/Syntax:** The Child Model relies on the explicit presence of quantized objects (determiners) to identify telicity. This is a high-precision but brittle rule (e.g., it fails on "I ate popcorn" vs

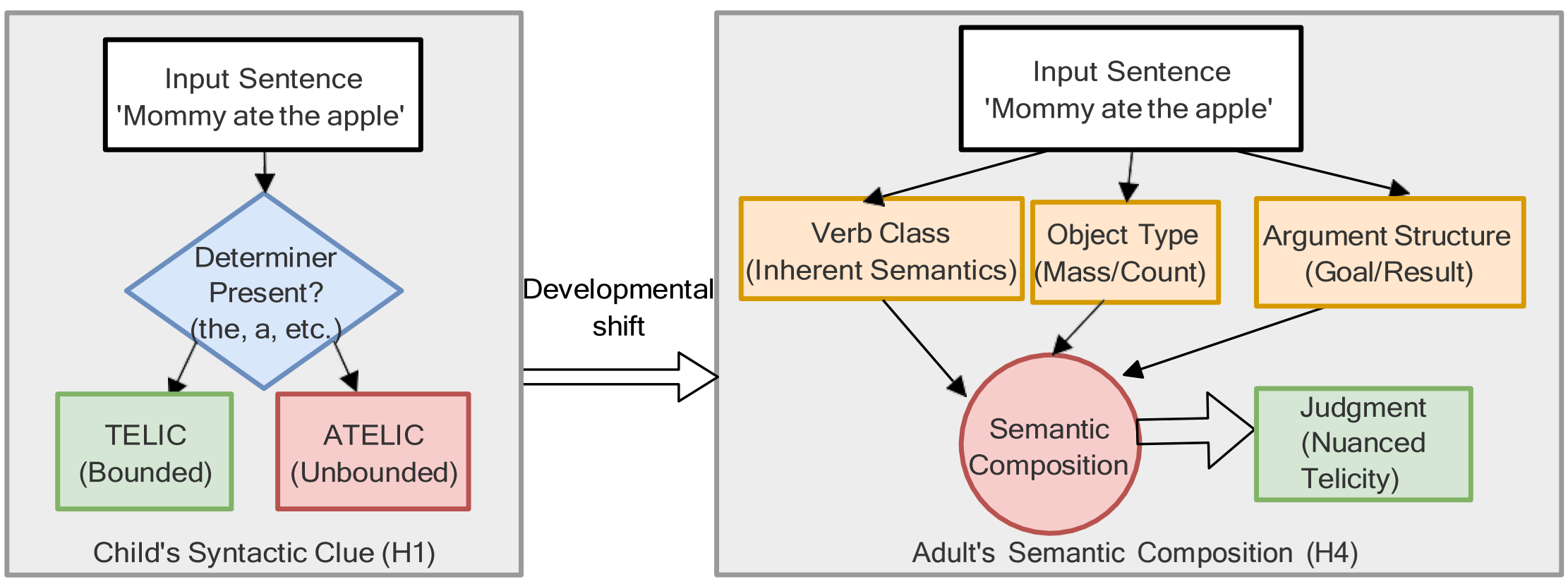

Figure 4: Conceptual model of the developmental shift in telicity acquisition

"I ate the popcorn" if the child doesn't understand mass nouns yet). It confirms that $H_1$ holds in child's production of telicity.

**Adult Model/Semantics:** The Adult Model integrates the inherent aspectual class of the verb itself (e.g., recognizing that *arrive* is inherently telic regardless of other cues). It confirms that $H_4$ holds in adult's production of telicity.

## 7 Conclusion

This study provides a novel computational bridge between formal event semantics and L1 acquisition theory. By connecting LLM-based surprisal with diagnostic probing, we have identified a developmental shift in the acquisition of telicity. Our results demonstrate that while children initially "bootstrap" event boundedness through a high-precision, surface-level "Determiner Clue" ($H_1$), adult speakers transition toward a more integrated model that prioritizes inherent lexical semantics ($H_4$). Future work will investigate whether this developmental pattern holds across morphologically richer (Spanish, German) or lesser (Mandarin, Thai) languages to determine if the "Determiner Clue" is a universal bootstrapping strategy or a language-specific adaptation to English grammar.

## Limitations

Our study utilizes GPT-2 trained on adult web text, which measures alignment with adult norms rather than with the language input or language production typical in child language acquisition. Furthermore, the reliance on manual CHILDES transcripts may overlook multi-modal cues, i.e., gestures.

## Ethical considerations

The human evaluation component of this study was conducted with the approval of the Michigan State University Institutional Review Board (IRB Protocol STUDY202500548). Three expert linguists in linguistics program at Michigan State University participated in the evaluation on a voluntary basis. All participants were provided with a detailed description of the task and research objectives, and written informed consent was obtained prior to data collection.

## Acknowledgment

We express our sincere gratitude to Dr. Jason Smith, Drake Howard, and Millie Hacker for their generous help with data annotation during Fall 2025 and Dr. Andrew McInnerney for his suggestion on coding. We also thank the audiences at the 39th Annual Conference on Careers, Alumni and Linguistics at Michigan State (CALMS) 2025, Human Sentence Processing (HSP39) and Midwest Speech and Language Days (MSLD) 2026 for their valuable feedback and comments.

# A Preprocessing and Annotation Statistics

Table 7 reports the size of the CHILDES subsets before and after preprocessing, along with the resulting telicity label distribution from the Difference in Surprisal method. Table 8 reports the label distribution of the 200-sentence expert-annotated sample used for validation, both per rater and against model predictions.

Table 7: Corpus size and label distribution before and after preprocessing.

| | **Child** | **Adult** |
|---|---|---|
| Raw utterances | 835,151 | 170,844 |
| Utterances after preprocessing | 417,181 | 106,388 |
| Telic (%) | 31% | 12% |
| Atelic (%) | 69% | 88% |

Table 8: Label distribution in the 200-sentence expert-annotated sample (by rater) vs. Model Prediction.

| | **Telic** | **Atelic** |
|---|---|---|
| Rater #1 | 45 | 155 |
| Rater #2 | 42 | 158 |
| Rater #3 | 46 | 154 |
| Model Prediction | 55 | 145 |